\documentclass[11pt]{amsart}
\usepackage{geometry}                
\usepackage{graphicx}
\usepackage{amssymb}
\usepackage{epstopdf}
\usepackage{xcolor}
\usepackage[utf8]{inputenc}
\usepackage{amssymb}
\usepackage{natbib}
\usepackage{hyperref}

\newcommand{\Remark}[1]{\ifodd\value{page} \normalmarginpar
 \else \reversemarginpar \fi \marginpar{{\footnotesize #1}} }
\newcommand{\RR}{\Bbb{R}}

\newcommand{\cling}{\mathbf{C}}                  
\newcommand{\Units}{\mathbf{U}}
\newcommand{\Ldrs} {{\mathcal T}}
\newcommand{\f}{\mathbf{f}}
\newcommand{\p}{\mathbf{p}}
\newcommand{\x}{\mathbf{x}}
\newcommand{\lead}{\mathbf{t}}
\newcommand{\ul}{\mathbf{u}}
\newcommand{\vl}{\mathbf{v}}
\newcommand{\zl}{\mathbf{z}}
\newcommand{\keyw}[1]{\textcolor{red}{\textit{#1}}}
\def\argmin{\mathop{\rm arg\ min}\nolimits}

\title{Clustering of Modal Valued Symbolic Data}
\author{Vladimir Batagelj \and Nataša Kejžar \and Simona Korenjak-Černe\\
University of Ljubljana, Slovenia}

\begin{document}
\maketitle

\begin{abstract}
{Symbolic Data Analysis is based on special descriptions of data -- symbolic objects (SO).
Such descriptions preserve more detailed information about units and their clusters than the
usual representations with mean values. A special kind of symbolic object is a
representation with frequency or probability distributions (modal values).
This representation enables us to consider in the clustering process the variables of
all measurement types at the same time.
In the paper a clustering criterion function for SOs is proposed such that the representative
of each cluster is again composed of distributions of variables' values over the cluster.
The corresponding leaders clustering method is based on this result. It is also shown that for
the corresponding agglomerative hierarchical method a generalized Ward's formula holds.
Both methods are compatible -- they are solving the same clustering optimization problem.\\
The leaders method efficiently solves clustering problems with large number of
units; while the agglomerative method can be applied alone on the smaller data set, or it
could be applied on leaders, obtained with compatible nonhierarchical clustering method.
Such a combination of two compatible methods enables us to decide upon the right number
of clusters on the basis of the corresponding dendrogram.\\
The proposed methods were applied on different data sets. In the paper, some
results of clustering of ESS data are presented.}
\end{abstract}

\section{Introduction}
\label{sec:1}

In traditional data analysis a unit is usually described with a list of (numerical, ordinal
or nominal) values of selected variables. In symbolic data analysis (SDA)
a unit of a data set can be represented, for each variable, with a more detailed description
than only a single value.
Such structured descriptions are usually called \keyw{symbolic objects} (SOs)
\cite{BockDid:2000}, \cite{SDA:2006}). A special type of symbolic objects are descriptions with
frequency or probability distributions.
In this way we can at the same time consider both --
a single value variables and variables with richer descriptions.
Computerization of data gathering worldwide caused the data sets getting huge. In order to be able to extract (explore) as much information as possible from such kind of data the predefined aggregation (preclustering) of the raw data is getting common. 

For example if a large store chain (that records each purchase its customers make) wants information about patterns of customer purchases, the very likely way would be to aggregate purchases of customers inside a selected time window. A variable for a customer can be a yearly shopping pattern on a selected item. Such a variable could be described with a single number (average yearly purchase) or with a symbolic description -- purchases on that item aggregated according to months. The second description is richer and allows for better analyses.

In order to retain and use more information about each unit during the clustering process,
we adapted two classical clustering methods:
\begin{itemize}
\item leaders method, a generalization of k-means method (\cite{Hartigan:1975},
 \cite{Anderberg:1973}, \cite{Diday:1979})
\item Ward's hierarchical clustering method (\cite{Ward:1963}).
\end{itemize}
{Both methods are \keyw{compatible} -- they are based on the same criterion function.
Therefore they are solving the same clustering optimization problem.
They can be used in combination: using the leaders method the size of the set
of units is reduced to a manageable number of leaders that can be further clustered
using the compatible hierarchical clustering method. It enables us to reveal the
relations among the clusters/leaders and also to decide, using the dendrogram,
upon the right number of final clusters.

Since clustering objects into similar groups plays an important role in the exploratory data analysis, many clustering approaches have been developed in SDA. Symbolic object can be compared using many different dissimilarities with different properties. Based on them many clustering approaches were developed. Review of them can be found in basic books and papers from the field: \cite{BockDid:2000}, \cite{BillDid:2003}, \cite{SDA:2006}, \cite{DidNoirh:2008}, and \cite{NoirhBrito:2011}.
Although most attention was given to clustering of interval data (de Carvalho, F.A.T.
and his collaborators) some methods were developed also for modal valued data that are close to our approach
(\cite{GowDid:1991}, \cite{IchYag:1994}, \cite{IFCS:1998}, \cite{VCL:2000}, \cite{IFCS:2002}, \cite{IrpVer:2006}, \cite{VerIrp:2010}}). 

The very recent paper of \cite{IrpVerCar:2014} describes the dynamic clustering approach to histogram data based on Wasserstein distance. This distance allows also for automatic computation of relevance weights for variables. The approach is very appealing but cannot be used when clustering general (not necessarily numerical) modal valued data.
In the paper \cite{CarSou:2010} the authors present an approach with dynamic clustering that could be (with a pre-processing step) used to cluster any type of symbolic data. For the dissimilarities adaptive squared Euclidean distance is used. One drawback to this approach is in the fact that when using dynamic clustering one has to determine the number of clusters in advance.
In the paper \cite{KimB:2011} the authors propose Ichino-Yaguchi dissimilarity measure extended to histogram data and in the paper \cite{KimB:2012} two measures (Ichino-Yaguchi and Gowda-Diday) extended to general modal valued data. They use these measures with divisive clustering algorithm and propose two cluster validity indexes that help one decide for the optimal number of final clusters. In the paper \cite{KimB:2013} even more general dissimilarity measures are proposed to use with mixed histogram, multi valued and interval data.

The aim of this paper is to provide a theoretical basis for a generalization of the compatible leaders and agglomerative hierarchical clustering methods for modal valued data with meaningful interpretations of clusters' leaders.
The novelty in our paper is in proposed additional dissimilarity measures (stemming directly from squared Euclidean distance) that allow the use of weights for each SO (or even its variable's component) in order to consider the size of each SO. It is shown that each of these dissimilarities can be used in a leaders and agglomerative hierarchical clustering method, thus allowing the user to chain both methods. In dealing with big data sets we can use leaders methods to shrink the big data set into a more manageable number of clusters (each represented by its leader) which can be further clustered via hierarchical method. Thus the number of final cluster is easily determined from the dendrogram.

When clustering units described with frequency distributions, the following problems can occur:
\begin{quote}\normalsize
\textbf{Problem 1:} The values in descriptions of different variables can be based on different number of original units. 
\end{quote}
A possible approach how to deal with this problem is presented in an application in \cite{TIMSS:2011}, where two related data sets (teachers and their students) are combined in an ego-centered network, which is presented with  symbolic data description.
\begin{quote}\normalsize
  \textbf{Problem 2:} The representative of a cluster is not a meaningful representative of the cluster.
\end{quote}
For example, this problem appears when clustering units are age-sex structures of the world's 
countries (e.g. \cite{IrpVer:2006, KB:2011}). In \cite{PP:2015} authors used a weighted agglomerative clustering approach, where clusters' representatives are real age-sex structures, for clustering population pyramids of the world's countries.
\begin{quote}\normalsize
  \textbf{Problem 3:} The squared Euclidean distance favors distribution components with the largest values.
\end{quote}

In clustering of citation patterns \cite{Kejzar:2011} showed that the selection of the squared Euclidean 
distance doesn't give very informative clustering results about citation patterns. The authors therefore 
suggested to use relative error measures. 

In this paper we show that all three problems can be solved using the generalized leaders and Ward's 
methods with an appropriate selection of dissimilarities and with an appropriate selection of weights. They produce more meaningful clusters' representatives. The paper also provides theoretical basis for compatible usage of both methods and extends methods with alternative dissimilarities (proposed in \cite{Kejzar:2011} for classical data representation) on modal valued SOs with general weights (not only cluster sizes).

In the following section we introduce the notation and the development of the adapted methods is presented. The third section describes an example analysis of the European Social Survey data set \cite{ESS:2010}. Section four concludes the paper.  In the  Appendix  we provide proofs that the alternative dissimilarities can also be used with the proposed approach.


\section{Clustering}
\label{sec:2}

The 
\textit{set of units} $\Units$ consists of symbolic objects (SOs). An SO $X$ is described with a
list of descriptions of variables $V_i, i=1,\dots, m$. In our general model, each variable is described with
a list of values $\f_{x_i}$
\[  x = [ \f_{x_1}, \f_{x_2}, \ldots, \f_{x_m} ],\]
where $m$ denotes the the number of variables and
\[  \f_{x_i} = [ f_{x_i1}, f_{x_i2}, \ldots, f_{x_ik_i}], \]
with $k_i$ being the number of terms (frequencies) $f_{x_ij}$ of a variable $V_i, i=1,\dots, m$.

Let $n_{x_i}$ be the count of values of a variable $V_i$
\[ n_{x_i} = \sum_{j=1}^{k_i} f_{x_ij} \]
then we get the corresponding probability distribution
\[
\p_{x_i} = \frac{1}{n_{x_i}} \f_{x_i}.
\]

In general, a frequency distribution can be represented as a vector or graphically as a barplot (histogram). To preserve the same description for the variables with different measurement scales, the range of the continuous variables or variables with large range has to be categorized (partitioned into classes).
In our model the values NA (not available) are treated as an additional category for each variable, but in some cases use of imputation methods for NAs would be a more recommended option.

Clustering data with leaders method or hierarchical clustering method are two approaches for
solving the clustering optimization problem. We are using the
criterion function of the following form
\[
 P(\cling) =   \sum_{C \in \cling}  p(C).   \label{eqP0}
\]
The \textit{total error} $P(\cling)$ of the clustering $\cling$ is the sum of
\textit{cluster errors} $p(C)$ of its clusters $C \in \cling$.

There are many possibilities how to express the cluster error $p(C)$. In this paper we
shall assume a model in which the error of a cluster is a sum of differences of its units
from the cluster's \keyw{representative} $T$.
For a given representative $T$ and a cluster $C$ we define
the cluster error with respect to  $T$:
\[
 p(C,T) =  \sum_{X \in C} d(X,T),   \label{eqpCT}
\]
where $d$ is a selected dissimilarity measure. The best representative $T_C$ is called
a \keyw{leader}
\[
 T_C =  \argmin_T p(C,T).   \label{eqTC}
\]
Then we define
 \begin{equation}
 p(C) =  p(C,T_C) = \min_T  \sum_{X \in C} d(X,T).  \label{eqp}
 \end{equation}

Assuming that the leader $T$  has the same structure of the description as SOs (i.e. it is represented with the list of nonnegative vectors $\lead_i$ of the size $k_i$ for each variable $V_i$). We do not require that they are distributions, therefore the representation space is $\Ldrs = (\RR_0^+)^{k_1} \times (\RR_0^+)^{k_2} \times \dots \times (\RR_0^+)^{k_m}$.

We introduce a dissimilarity measure between SOs and $T$ with
\[
 d(X,T) =   \sum_i  \alpha_i  d_i(X,T), \quad \alpha_i \geq 0, \quad
 \sum_i  \alpha_i = 1,  \label{eqdiss}
\]
where $\alpha_i$ are weights for variables (i.e. to be able to determine a more/less important variables) and
\[
  d_i(X,T) =    \sum_{j=1}^{k_i} w_{x_ij} \delta(p_{x_ij},t_{ij}),
 \quad w_{x_ij} \geq 0  \label{eqdg}
\]
where $w_{x_ij}$ are weights for each variable's component.
This is a kind of a generalization of the \keyw{squared Euclidean distance}. 
Using an alternative \keyw{basic dissimilarity} $\delta$ we can address the \keyw{problem 3}. Some examples of basic dissimilarities $\delta$ are presented in Table~\ref{t:relative}.
It lists the basic dissimilarities between the unit's component and the leader's component that were proposed in \cite{Kejzar:2011} for classical data representation. In this paper we extend them to modal valued SOs.

The weight $w_{x_ij}$
can be for the same unit $X$ different for each variable $V_i$ and also for each of its components. 
With weights we can include in the clustering process different number of original units for each variable (solving \keyw{problem 1} and/or \keyw{problem 2}) and they also allow a regulation of importance of each variable's category.
For example, the population pyramid of a country $X$ can be represented with two symbolic variables
(one for each gender), where people of each gender are represented with the distribution
over age groups. Here, $w_{x_1}$ is the number of all men and $w_{x_2}$ is
the number of all women in the country $X$.

To include and preserve the information about the variable distributions and their size
throughout the clustering process (\keyw{problem 2}), the following has to hold when merging two disjoint
clusters $C_u$ and $C_v$ (a cluster may consist of one unit only):
\[
\p_{(uv)_i} = \frac{1}{w_{(uv)_i}}\f_{(uv)_i}
\]
where $\p_{(uv)_i}$ denotes the relative distribution of the variable $V_i$ of the joint cluster $C_{(uv)} = C_u \cup C_v,\ \f_{(uv)_i}$ the frequency distribution of variable $V_i$ in the joint cluster and $w_{(uv)_i}$ the weight (count of values) for that variable in the joint cluster. 

Although we are using the notation $\f$ which is usually used for frequencies, other interpretations of $\f$ and $w$ are possible. For example $w$ is the money spent, and $\f$ is the distribution of the money spent on a selected item in a given time period. 

\subsection{Leaders method}

Leaders method, also called dynamic clouds method \citep{Diday:1979},
is a generalization of a popular nonhierarchical clustering k-means method
\citep{Anderberg:1973,Hartigan:1975}.
The idea is to get the "optimal" clustering into a
pre-specified number of clusters with an iterative procedure.
For a current clustering the leaders are determined as the best representatives of its clusters; and the new
clustering is determined by assigning each unit to the nearest leader.
The process stops when the result stabilizes.

In the generalized approach, two steps should be elaborated:
\begin{itemize}
  \item how to determine the new leaders;
  \item how to determine the new clusters according to the new leaders.
\end{itemize}

\subsubsection{Determining the new leaders}

Given a cluster $C$, the corresponding leader $T_C \in \Ldrs$ is the solution of the
problem (Eq. \ref{eqp})
\[ T_C = \argmin_T  \sum_{X \in C} d(X,T) =
\argmin_T  \sum_{X \in C} \sum_{i}  \alpha_i  d_i(X,T)  \]
\[ = \argmin_T \sum_{i} \alpha_i \sum_{X \in C} d_i(X,T)  =
\big[ \argmin_{\lead_i} \sum_{X \in C} d_i(X,T)  \big]_{i=1}^m
\]
Denoting $T_C =  [ \lead^*_1, \lead^*_2, \ldots , \lead^*_m ]$, where $\lead^*_i \in (\RR_0^+)^{k_i}, i=1,2, \ldots, m$, we get the following requirement: $\lead^*_i = \argmin_{\lead_i} \sum_{X \in C} d(\x_i,\lead_i)$.

Because of the additivity of the model we can observe each variable separately and simplify the notation by omitting the index $i$.
\[ \lead^* = \argmin_{\lead} \sum_{X \in C} d(\x,\lead) =
   \argmin_{\lead} \sum_{X \in C} \sum_{j=1}^{k} w_{xj} \delta(p_{xj},t_j) \]
\[ = \argmin_{\lead} \sum_{j=1}^{k} \sum_{X \in C} w_{xj} \delta(p_{xj},t_j)
   = \big[ \argmin_{t_j \in \RR} \sum_{X \in C} w_{xj} \delta(p_{xj},t_j) \big]_{j=1}^k \]
Since in our model also the components are independent we can optimize component-wise and omit the index $j$
\begin{equation} \label{leaderopt}
  t^* = \argmin_{t \in \RR} \sum_{X \in C} w_{x} \delta(p_{x},t) 
\end{equation}
$t^*$ is a kind of Fr\'{e}chet mean (median) for a selected basic dissimilarity $\delta$.

This is a standard optimization problem with one real variable. The solution
has to satisfy the condition 
\begin{equation}
\frac{\partial}{\partial t} \sum_{X \in C} w_{x} \delta(p_{x},t) = 0
 \label{leadereq}
\end{equation}

\paragraph{{Leaders for $\delta_1$}}

Here we present the derivation only for the basic dissimilarity $\delta_1$. The derivations for other dissimilarities $\delta$ from Table~\ref{t:relative} are given in the appendix.

\begin{table}[htb!]
\begin{center}
\caption{\label{t:relative}The basic dissimilarities and
the corresponding cluster leader, the leader of the merged clusters and dissimilarity between merged clusters. Indices $i$ and $j$ are omitted.}

\parbox{85mm}{\strut\hspace*{5mm}\\
$\begin{array}{ccccc}
\hline\noalign{\smallskip}
   & \delta(x,t) & t_C^*  & z & D(C_u,C_v) \\ \noalign{\smallskip}\hline\noalign{\smallskip}
 \delta_1 & (p_{x}-t)^2  & \frac{P_C}{w_C} & \displaystyle \frac{w_{u} u + w_{v} v}{w_{u} + w_{v}} & \frac{w_{u}\cdot w_{v}}{w_{u} + w_{v}}(u-v)^2
   \\ 
 \delta_2 & (\frac{p_{x}-t}{t})^2 & \frac{Q_C}{P_C} & \displaystyle \frac{u P_{u}+v P_{v}}{P_{u}+P_{v}} & \frac{P_{u}}{u} (\frac{u-z}{z})^2 + \frac{P_{v}}{v} (\frac{v-z}{z})^2
   \\ 
 \delta_3 & \frac{(p_{x}-t)^2}{t} & \sqrt{\frac{Q_C}{w_C}} & \displaystyle \sqrt{\frac{u^2 w_{u}+v^2 w_{v}}{w_{u} + w_{v}}} & w_u \frac{(u-z)^2}{z} + w_v \frac{(v-z)^2}{z}
   \\ 
 \delta_4 & (\frac{p_{x}-t}{p_{x}})^2 & \frac{H_C}{G_C} & \displaystyle \frac{H_{u}+H_{v}}{\frac{H_{u}}{u} + \frac{H_{v}}{v}} & G_u (u-z)^2 + G_v (v-z)^2
   \\ 
 \delta_5 & \frac{(p_{x}-t)^2}{p_{x}}  & \frac{w_C}{H_C} & \displaystyle \frac{w_{u}+w_{v}}{H_{u}+H_{v}} & w_u \frac{(u-z)^2}{u} + w_v \frac{(v-z)^2}{v}
   \\ 
 \delta_6 & \frac{(p_{x}-t)^2}{p_{x}t} & \sqrt{\frac{P_C}{H_C}} & \displaystyle \sqrt{\frac{P_{u}+P_{v}}{\frac{P_{u}}{u^2} + \frac{P_{v}}{v^2}}} & \frac{P_{u}}{u} \frac{(u-z)^2}{uz} + \frac{P_{v}}{v} \frac{(v-z)^2}{vz}
   \\ 
\noalign{\smallskip}\hline
\end{array}$
}
\parbox{18mm}{
\begin{eqnarray*} \label{abbre}
w_C &=& \sum_{X \in C} w_{x} \\
P_C &=& \sum_{X \in C} w_{x} p_{x}\\
Q_C &=& \sum_{X \in C} w_{x} p^2_{x}\\
H_C &=& \sum_{X \in C} \frac{w_{x}}{p_{x}}\\
G_C &=& \sum_{X \in C} \frac{w_{x}}{p^2_{x}}
\end{eqnarray*}
}
\end{center}

\end{table}

The traditional clustering criterion function in $k$-means and Ward's
clustering methods is based on the squared Euclidean distance dissimilarity $d$ that is
based on the basic dissimilarity
 $\delta_1(p_x,t) = (p_x - t)^2$. For it we get from~(\ref{leadereq})
\[ 0 = \sum_{X \in C} w_{x} \frac{\partial}{\partial t} (p_x - t)^2 =
 -2 \sum_{X \in C} w_{x} (p_x - t) \]
Therefore
\begin{equation}
t^* = \frac{\sum_{X \in C} w_x p_x}{\sum_{X \in C} w_x} = \frac{P_C}{w_C} \label{T2}
\end{equation}

The usage of selected weights in the dissimilarity $\delta_1$ provides meaningful cluster representations, resulting from the following two properties:

\paragraph{{Property 1}}

Let $w_{x_ij} = w_{x_i}$ then for each $i = 1,\ldots,m$:
\[ \sum_{j=1}^{k_i} t^*_{ij} =
   \frac{1}{w_C} \sum_{j=1}^{k_i} \sum_{X \in C} w_{x_i} p_{x_ij}
   =  \frac{1}{w_C}  \sum_{X \in C} w_{x_i} \sum_{j=1}^{k_i} p_{x_ij} =
  \frac{1}{w_C}  \sum_{X \in C} w_{x_i} = 1 \]
If the weight $w_{x_ij}$ is the same for all components of variable $V_i$,
$w_{x_ij} = w_{x_i}$, then for $\delta_1$ the leaders' vectors $t^*_i$ are \keyw{distributions}.

\paragraph{{Property 2}}

Let further $w_{x_ij} = n_{x_i}$ then for each cluster C, $i = 1,\ldots,m$ and
$j = 1, \ldots,k_i$:

\[ t^*_{Cij} = \frac{\sum_{X \in C} n_{x_i} p_{x_ij}}{\sum_{X \in C} n_{x_i}}
   = \frac{\sum_{X \in C} f_{x_ij}}{\sum_{X \in C} n_{x_i}}
   = \frac{f_{Cij}}{n_{Ci}} = p_{Cij} \]
Note that in this case the weight $w_{x_ij}$ is constant for all components
of the same variable.
This result provides a solution to the \keyw{problem 2}.\medskip


For each basic dissimilarity $\delta$ the corresponding optimal leader, the leader of the merged clusters, and the dissimilarity $D$ between clusters are given in Table~\ref{t:relative}. 

\subsubsection{Determining new clusters}

Given leaders $\mathbf{T}$ the corresponding optimal clustering $\cling^*$ is determined from
\begin{equation}
 P(\cling^*) = \sum_{X \in \Units} \min_{T \in \mathbf{T}} d(X,T)  =
  \sum_{X \in \Units} d(X,T_{c^*(X)}),  \label{eqPC}
 \end{equation}
where
$ c^*(X) = \argmin_k d(X,T_k) $.
We assign each unit $X$ to the closest leader $T_k \in \mathbf{T}$.


In the case that some cluster becomes empty, usually the most distant unit from some
other cluster is assigned to it.
In the current version of R package \textbf{clamix} \citep{Clamix} the most dissimilar unit from all the cluster
leaders is assigned to the empty cluster.

\subsection{Hierarchical method}

The idea of the agglomerative hierarchical clustering procedure is a step-by-step merging
of the two closest clusters starting from the clustering in which each unit forms its own
cluster. The computation of dissimilarities between the new
(merged) cluster and the remaining other clusters has to be specified.

\subsubsection{Dissimilarity between clusters}
To obtain the compatibility with the adapted leaders method, we define the dissimilarity
between clusters $C_u$ and $C_v$, $C_u \cap C_v = \emptyset$, as
\citep{Batagelj:1988}
\begin{equation} \label{eqDuv}
   D(C_u,C_v) = p(C_u \cup C_v) - p(C_u) - p(C_v).
\end{equation}

Let us first do some general computation.
$\ul_i$ and $\vl_i$ are $i$-th variables of the leaders $U$ and $V$ of clusters $C_u$ and $C_v$,
and $\zl_i$ is a component of the leader $Z$ of the cluster $C_u \cup C_v$. Then

\[ D(C_u,C_v) = p(C_u \cup C_v) - p(C_u) - p(C_v) = \]
\[ = \sum_i \alpha_i \left[ \sum_{X \in C_u \cup C_v} d_i(X,Z)
 - \sum_{X \in C_u} d_i(X,U)
 - \sum_{X \in C_v} d_i(X,V) \right]
 = \sum_i \alpha_i D_i(C_u,C_v) \]
Since $C_u \cap C_v = \emptyset$  we have
\begin{equation}\label{Suv}
  D_i(C_u,C_v)
= \sum_{X \in C_u} \left[d_i(X,Z) - d_i(X,U)\right] +
    \sum_{X \in C_v} \left[d_i(X,Z) - d_i(X,V)\right]  
      = S_{ui} + S_{vi} 
\end{equation}
Let us expand the first term
\begin{equation}\label{Su}
 S_{ui}  = \sum_{X \in C_u}  \sum_j w_{x_ij}
 \left[\delta(p_{x_ij},z_{ij}) - \delta(p_{x_ij},u_{ij})\right] = \sum_j S_{uij}
\end{equation}

\subsubsection{Generalized Ward's relation for $\delta_1$}

Now we consider a selected basic dissimilarity 
$\delta_1(p_x,t) = (p_x - t)^2$. We get (omitting $ij$-s)
\[ S_{uij} = \sum_{X \in C_u}  w_{x} \left[(p_{x}-z)^2 - (p_{x}-u)^2\right]
   = \sum_{X \in C_u}  w_{x} (z^2 - 2p_{x} z + 2 p_{x} u -u^2) = \]
as we know (\ref{T2}): $P_{u} = w_{u} u$
\[ =  w_{u} z^2 - 2 w_{u} u ( z - u) - w_{u} u^2 =
    w_{u} (z - u)^2. \]
Therefore
\[ D_{ij}(C_u,C_v) =  w_{u} (z-u)^2 + w_{v} (z-v)^2 \]
\[ = w_{u} (z^2 - 2u z + u^2)
               + w_{v} (z^2 - 2v z + v^2) \]
\[ = z^2 (w_{u}+w_{v}) - 2z (w_{u} u
               + w_{v} v) + w_{u}u^2 + w_{v} v^2 . \]

We can express the new cluster leader's element $z$ also in a different way.
\[ w_{z} z = P_{z} = \sum_{X \in C_u \cup C_v} w_{x} p_{x}
   = \sum_{X \in C_u} w_{x} p_{x} + \sum_{X \in C_v} w_{x} p_{x} =
    w_{u} u + w_{v} v \]
Therefore
\[  z  = \frac{w_{u} u + w_{v} v}{w_{u}+w_{v}}. \]
This relation is used in the expression for $D_{ij}(C_u,C_v)$:
\[
D_{ij}(C_u,C_v) = w_u u^2 + w_v v^2 - (w_u+w_v) z^2 \]
\[ = w_u u^2 + w_v v^2 - (w_u+w_v)(\frac{w_u u +
     w_v v}{w_u+w_v})^2 \]
\[ =  \frac{w_u \cdot w_v}{w_u+w_v}(u-v)^2 .\]
and finally, reintroducing $i$ and $j$, we get
\begin{equation}\label{wardrel}
D(C_u,C_v) = \sum_i \alpha_i \sum_j \frac{w_{uij}\cdot w_{vij}}{w_{uij}+w_{vij}}(u_{ij}-v_{ij})^2
\end{equation}
a \keyw{generalized Ward's relation}.
Note that this relations holds also for singletons $C_u=\{X\}$ or $C_v=\{Y\}$,\  $X,Y \in \Units$.

\paragraph{{Special cases of the generalized Ward's relation}}

When $w_{x_i} = w_x$ is the same for all variables
$V_i, i=1,\ldots,m$, the Ward's relation (\ref{wardrel}) can be simplified.
In this case we have
\[ \lead_{i}^* = \frac{1}{w_C} \sum_{X \in C} w_{x}\cdot \p_{x_i}, \]
where the sum of weights $w_C = \sum_{X \in C} w_{x_i} = \sum_{X \in C} w_{x}$
\ is independent of $i$.
Therefore we have
\[ D(C_u,C_v) = \frac{w_u\cdot w_v}{w_u+w_v} \sum_i \alpha_i  (\ul_i-\vl_i)^2 =
\frac{w_u\cdot w_v}{w_u+w_v} d(\ul,\vl), \]

In the case when for each variable $V_i$ all $w_{x_i} = 1$, further simplifications are possible.
Since $\sum_{X \in C} 1 = |C|$ we get
\[ \lead_{i}^* = \frac{1}{|C|} \sum_{X \in C} \p_{x_i} \]
and 
\[ D(C_u,C_v) = \frac{|C_u|\cdot |C_v|}{|C_u|+|C_v|} \sum_i \alpha_i  (\ul_i-\vl_i)^2 =
\frac{|C_u|\cdot |C_v|}{|C_u|+|C_v|} d(\ul,\vl). \]


\subsection{Huygens Theorem for $\delta_1$}
\label{sec:HT}

Huygens theorem has a very important role in many fields. In statistics it can be related to the decomposition of sum of squares, on which the analysis of variance is based. 
In clustering it is commonly used for deriving clustering criteria.
It has the form
\begin{equation}
   TI = WI + BI,
\end{equation}
where $TI$ is the \keyw{total inertia}, $WI$ is the \keyw{inertia within clusters} and $BI$ is the \keyw{inertia between clusters}.

Let $\lead_\Units$ denote the leader of the cluster consisting of all units 
$\Units$. Then we define \citep{Batagelj:1988}
\begin{eqnarray*}
 TI & = & \sum_{X \in \Units} d(X,\lead_\Units) \\
 WI & = & P(\cling)  = \sum_{C \in \cling}\sum_{X \in C} d(X,\lead_C) \\
 BI & = & \sum_{C \in \cling} d(\lead_C,\lead_\Units)
\end{eqnarray*}

For a selected dissimilarity $d$ and a given set of units $\Units$ the value
of total inertia $TI$ is fixed. Therefore, 
if Huygens theorem holds, the minimization of the within inertia $WI= P(\cling)$ is equivalent to the maximization of the between inertia $BI$.

To prove that Huygens theorem holds for $\delta_1$ we proceed as follows.
Because of the additivity of $TI$, $WI$ and $BI$ and the component-wise definition
of the dissimilarity $d$, the derivation can be limited only to a single variable
and a single component. The subscripts $i$ and $j$ are omitted. We have
\begin{eqnarray*}
TI - WI & = & \sum_{C \in \cling} \sum_{X \in C} w_{x} \left [  (p_x - t_\Units)^2 - (p_x - t_C)^2  \right ]   \\
            & = & \sum_{C \in \cling} \sum_{X \in C} w_{x} \left (  -2p_x \ t_\Units + t_\Units^2 + 2p_x \ t_C + t_C^2   \right )\\
            & =  & \sum_{C \in \cling} \left ( -2 w_C \  t_C \  t_\Units + w_C \  t_\Units^2 + 2 w_C \  t_C^2 + w_C \  t_C^2  \right ) \\ 
          & = & \sum_{C \in \cling} w_C (t_C - t_\Units)^2 = \sum_{C \in \cling} d(t_C,t_\Units) = BI 
\end{eqnarray*}
This proves the theorem. In the transition from the second line to the third line
we considered that for $\delta_1$ holds (Eq. (\ref{T2}))
 $\sum_{X \in C} w_{x} p_x = w_C t_C$.


\section{Example}
 
The proposed methods were successfully applied on  different data sets: population pyramids, TIMSS, cars, foods, citation patterns of patents, and others. 
To demonstrate some of the possible usages of the described methods, some results of clustering of selected subset of the European Social Survey data set are presented.

The data set \cite{ESS:2010} is an output from an academically-driven social survey. Its main purpose
is to gain insight into behavior patterns, belief and attitudes of Europe's populations \citep{ESS:www}. The survey covers over 30 nations
and is conducted biennially. The survey data for the Round 5 (conducted in 2010) consist of 662 variables and include more than 50,000 respondents.
For our purposes we focused on the variables that describe household structure: (a) the gender of person in household, (b) the relationship to respondent in household (c) the year of birth of person in household  and (d) the country of residence for respondent, therefore also the country of the household. From these variables (the respondent answered the first three questions for every member of his/her household) symbolic variables (with counts of household members) were constructed.
Variable $V_3$ was the only numeric variable and therefore a decision had to be made of how to choose the category borders. 
From the economic point of view categorization into economic groups of working population is the most meaningful, therefore we chose this categorization in the first variant ($V_{3a}$ according to working population denoted with WP). But since demographic data about age are usually aggregated into five-year or ten-year groups, we also consider ten-year intervals as a second option of a categorization of the age variable ($V_{3b}$ in 10-year intervals denoted with AG).  
\begin{itemize}
\item ${V_1}$: gender (2 components):\\
$\{male: f_{11},\;female: f_{12}\}$
\item ${V_2}$: categories of household members (7 components, respondent constantly 1): \\
$\{respondent:f_{21}=1,\;partner:f_{22},\;offspring:f_{23},\;parents:f_{24},\;siblings:f_{25},\;relatives:f_{26},\;others:f_{27}\}$
\item ${V_3}$: year of birth for every household member:
\begin{itemize}
\item ${V_{3a}}$: according to working population (5 components):\\
$\{0-19\;years:f_{31},\;20-34\;years:f_{32},\;35-64\;years:f_{33},\;65+\;years:f_{34},\; NA:f_{35}\}$ 
\end{itemize}
or
\begin{itemize}
\item ${V_{3b}}$: 10-year groups (10 components):\\
$\{0-9,\;10-19,\;20-29,\;30-39,\;40-49,\;50-59,\;60-69,\;70-79,\;80+,\;NA\}$
\end{itemize}
\item ${V_4}$: country of residence (26 components, all but one with value zero):\\
$\{Belgium: f_{41}, \ldots, Ukraine: f_{4,26} \}$	
\end{itemize}

There were 641 respondents with missing values at year of birth therefore it seemed reasonable to add the category NA to variable $V_3$. 
That variant of handling missing values is very naive and could possibly lead to biased results (i.e. it could be conjectured that birth years of very old or non-related family members are mostly missing so they could form a special pattern). A refined clustering analysis would better use one of the well known imputation methods (e.g. multiple imputation, \cite{Rubin:1987}).
Note that for each unit (respondent) in the data set the components of variables $V_1$ to $V_3$ sum into a constant number (the number of all household members of that respondent).
For the last variable $V_4$, the sum equals 1. 

Design and population weights are supplied by the data set for each unit --- respondent. In order to get results that are representative of the EU population, both weights should be used also for households. Because special weights for households are not available, we used weights provided in data set in our demonstration: each unit's (respondent's) symbolic variables were before clustering multiplied by design and population weight. $w_{V_i}$ (used in the clustering process) for variables $V_1$ to $V_3$ was then the number of household members multiplied by both supplied weights and for $V_4$ the product of both weights alone.

\subsection{Questions about household structures}

Our motivation for clustering this data set was the question what are the main European household patterns? And further, does the categorization of ages of people from households influence the outcome of best clustering results a lot? To answer this part, two data sets with different variable sets were constructed: (a) data set with ages according to working population (further denoted as variable set WP) and (b) with ages split in nine 10-year groups (further denoted as variable set AG). Clustering was done on the three variables (gender, category of household members and  age-groups of household members). 

Since we were interested also whether the household patterns differ according to countries, we inspected variable 'country' after clustering. However this does not answer the following question \textit{Does the country influence the household patterns?} To be able to say something about that, 'country' has to be included in the data set and a third clustering on data with all four variables was performed. 

\subsection{Clustering process}

The set of units is relatively large --- 50,372 units. Therefore the clustering had to be done in two steps:
\begin{enumerate}
  \item cluster units with non-hierarchical leaders method to get smaller (20) number of clusters with their leaders;
  \item cluster clusters from first point (i.e. leaders) using hierarchical method to get a small number of final clusters.
\end{enumerate}
The methods are based on the same criterion function (minimization of the cluster errors based on the generalized squared Euclidean distance with $\delta = \delta_1$).

10 runs of leaders clustering were run for each data set (two data sets with three variables and one with four variables (including country)). The best result (20 leaders) of them for each data set was further clustered with hierarchical clustering. The dendrogram and final clusters were visually evaluated. 
Where variable 'country' was not included in the data set it was plotted later for each of the four groups and its pattern was examined.
Generally more runs of leaders algorithm are recommended. 
Since this example serves as an illustrative case and in ten runs the result was shown to be very stable we used ten runs only, but in actual application more runs of the leaders method would be recommended. 
The number of final clusters was selected with eyeballing the dendrogram selecting to cut where dissimilarity among clusters had the highest jump (apart from clustering in only two groups).

\subsection{Results}

\begin{figure}
  \includegraphics[width=\textwidth]{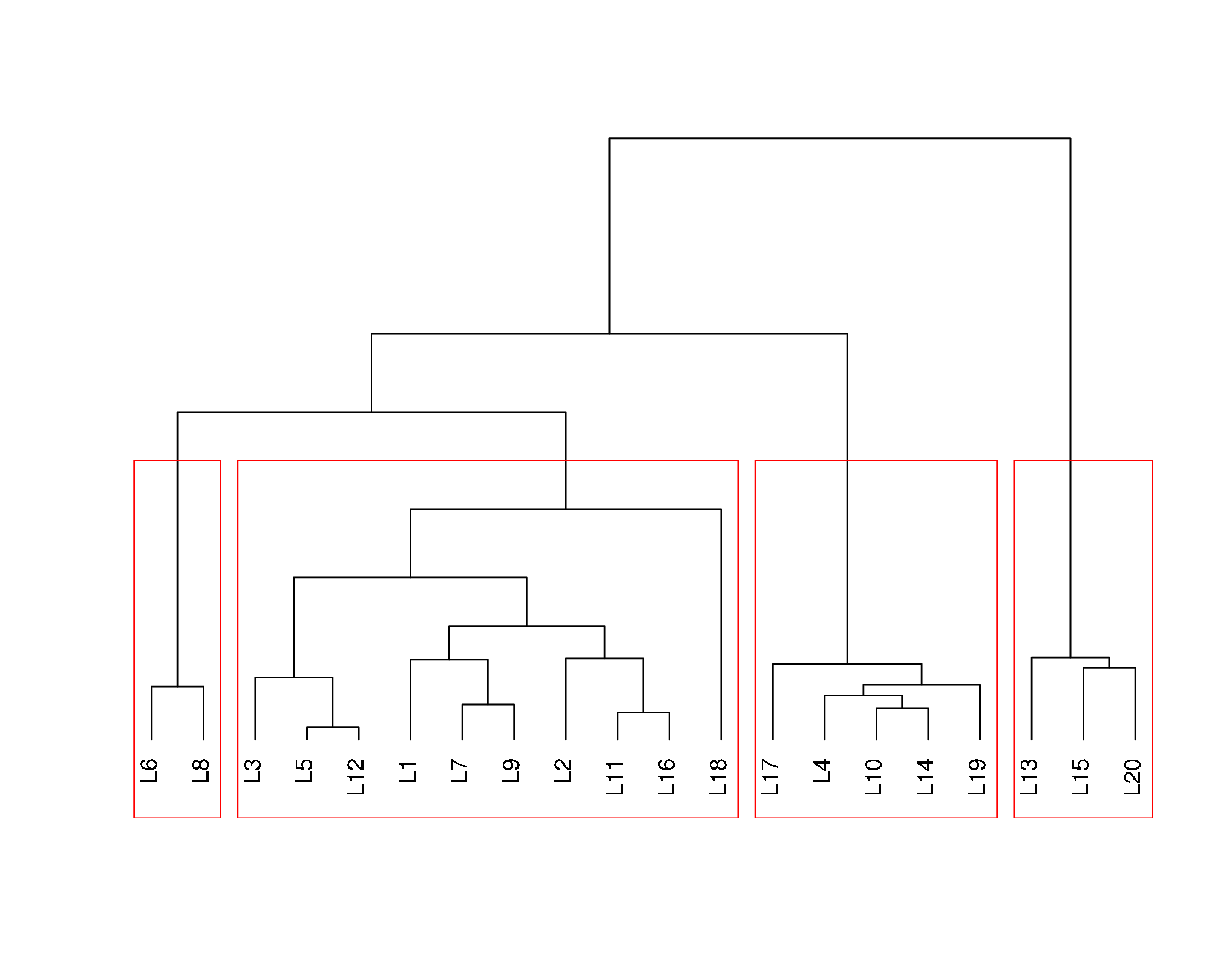}
\caption{Dendrogram for best leaders clustering for variable set WP with 
working population age groups.}\label{f: dendro RGY4}       
\end{figure}

\begin{figure}
  \includegraphics[width=\textwidth]{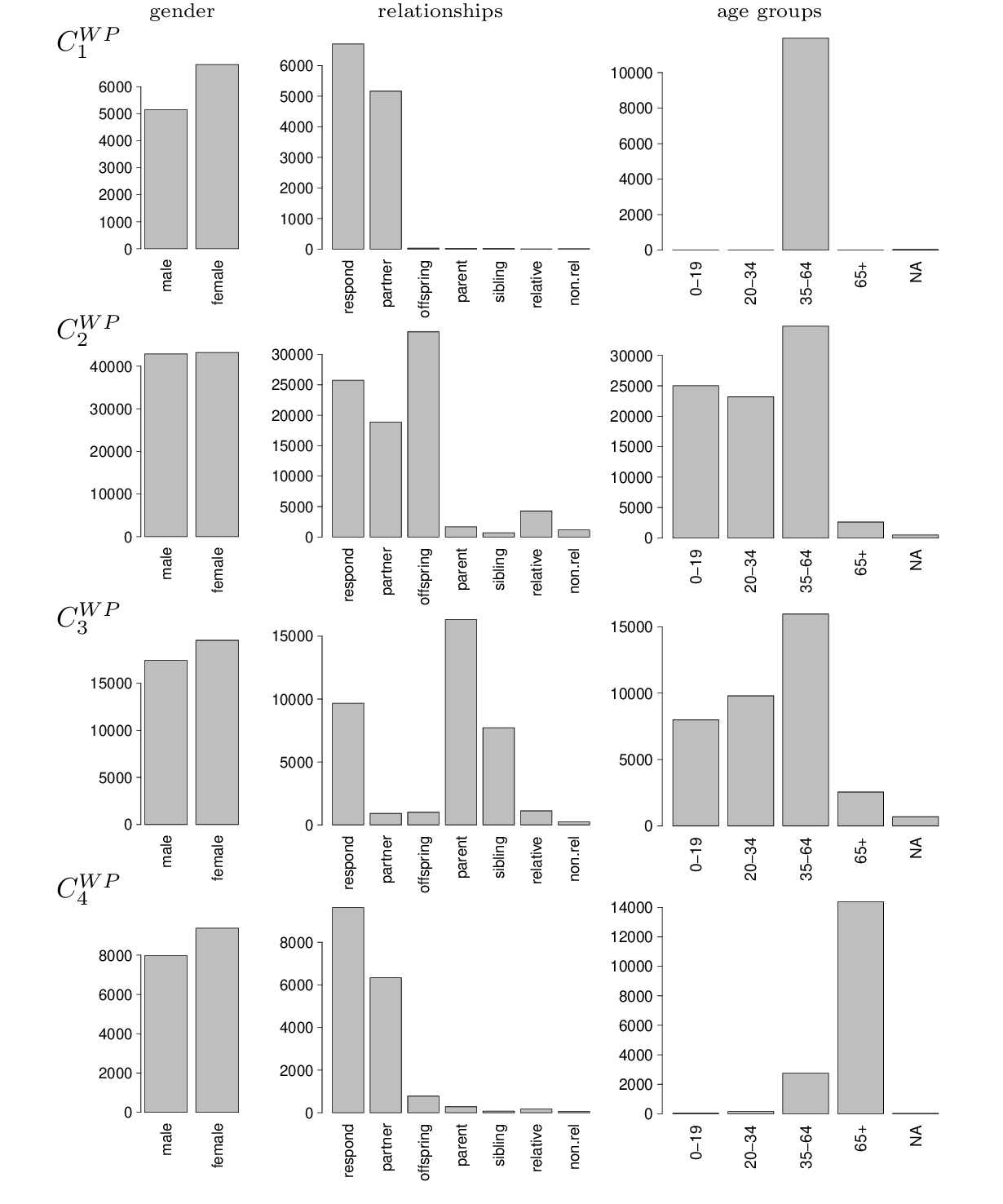}
\caption{Variable distributions for final 4 clusters ($C^{WP}_1$ with 7,134, $C^{WP}_2$ with 24,049, $C^{WP}_3$ with 7,280, and $C^{WP}_4$ with 11,909 households) 
with working population age groups.}\label{f: var RGY4}
\end{figure}

\begin{figure}
  \includegraphics[width=\textwidth]{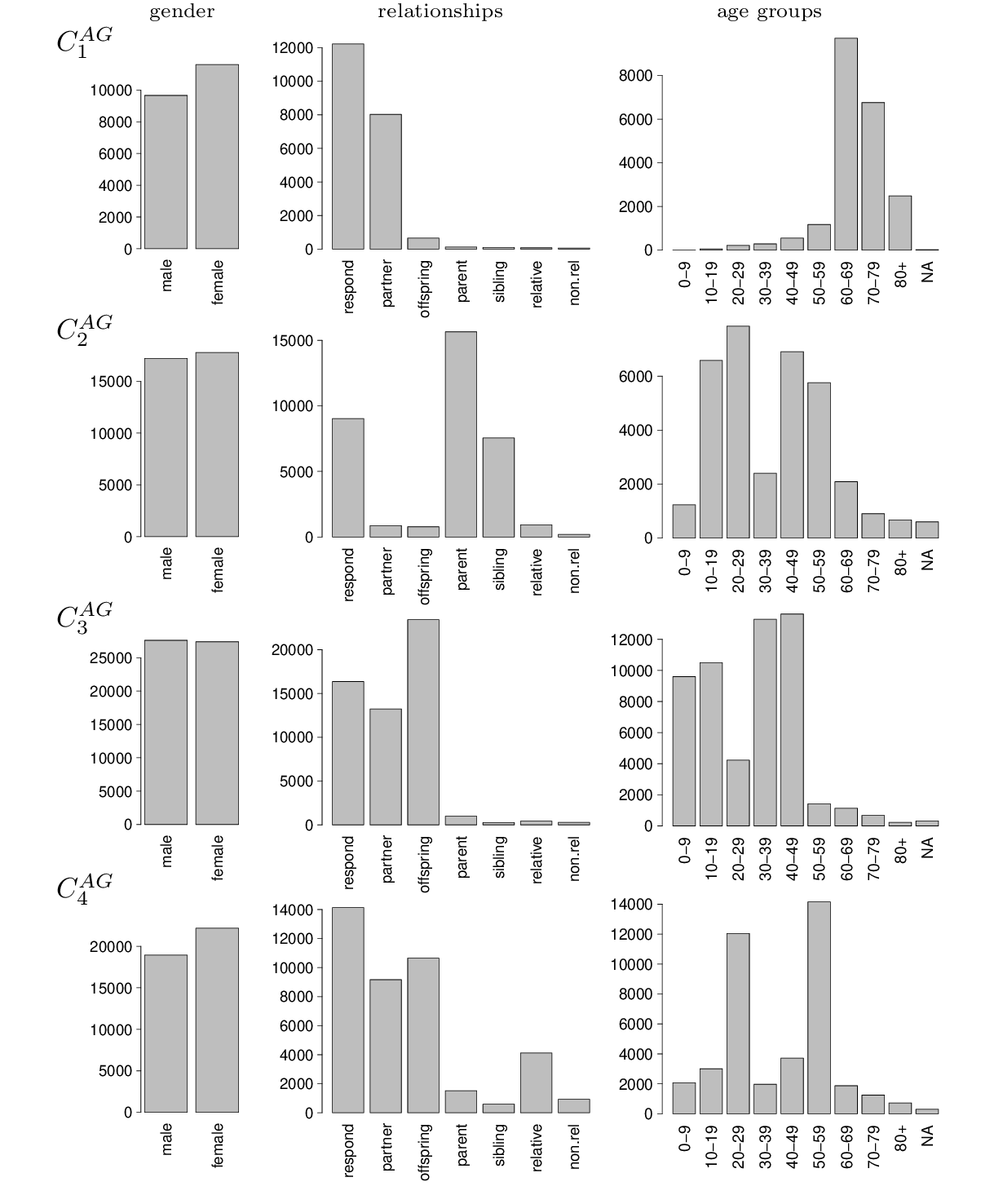}
\caption{Variable distributions for final 4 clusters for variable set AG ($C^{AG}_1$ with 14,975, $C^{AG}_2$ with 6,939, $C^{AG}_3$ with 15,800, and $C^{AG}_4$ with 12,658 households) with 10-year age categories.}\label{f: var RGY9}
\end{figure}

\begin{figure}
  \includegraphics[width=\textwidth]{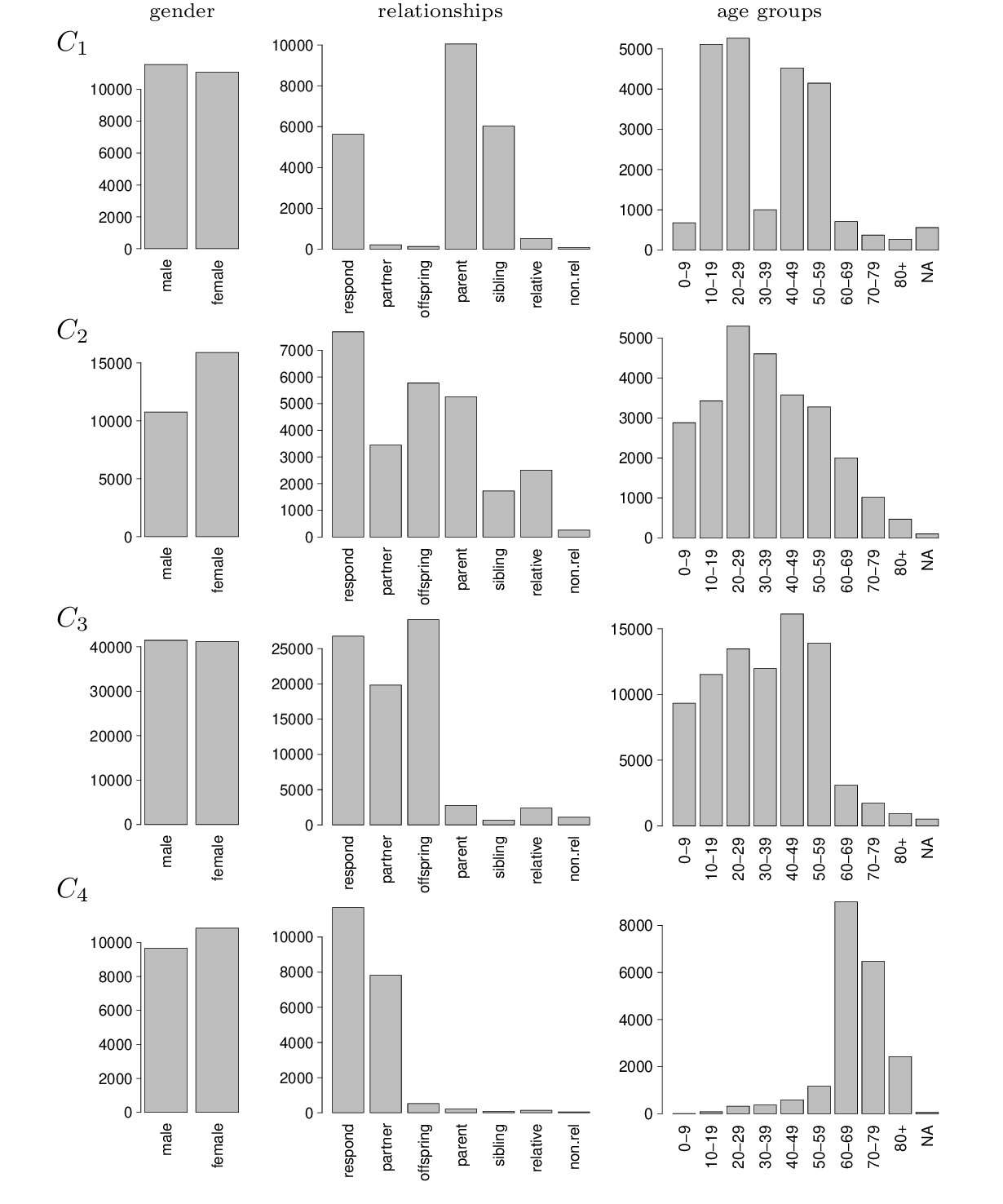}
\caption{Variable distributions for final 4 clusters for variable set Co with 10-year age categories.}\label{f: var RGYC9}
\end{figure}

The results were very stable. In Figure~\ref{f: dendro RGY4}  the dendrogram on the best leaders (with minimal leaders criterion function) for the variable set WP with ages according to working population  $\{V_1,V_2,V_{3a}\}$ is presented. For other two clusterings, i.i.e.e. for the variable set AG with 10-year age groups  $\{V_1,V_2,V_{3b}\}$ and the variable set Co with 'country' included $\{V_1,V_2,V_{3b},V_4\}$ the dendrograms look similar and are not displayed. The variable distributions of the final clusters are presented on Figure \ref{f: var RGY4} for the variable set WP, on Figure \ref{f: var RGY9} for the variable set AG, and on Figure \ref{f: var RGYC9} for the variable set with Co.

There is one large ($C^{WP}_2$ with 24,049 households), one middle sized ($C^{WP}_4$ with 11,909 households) and two small clusters ($C^{WP}_1$ with 7,134 and $C^{WP}_3$ with 7,280 households) in the result for the variable set WP.  For the variable set AG, three relatively medium size clusters ($C^{AG}_4$ with 12,658, $C^{AG}_1$ with 14,975, and $C^{AG}_3$ with 15,800 households) and a small one ($C^{AG}_2$ with 6,939 households) were detected.

Inspecting variable distributions, one can see (as expected) that gender is not a significant separator variable. The other two variables however both reveal household patterns. From Figure~\ref{f: var RGY4} and Figure~\ref{f: var RGY9} we see that most of the patterns 
can be matched between the clustering results with WP (working population age groups) and AG (10-year age groups): 
$C^{WP}_3$ with $C^{AG}_2$;  
$C^{WP}_2$ with $C^{AG}_3$; cluster $C^{AG}_1$ is split into two clusters in the WP clustering $C^{WP}_1$ and $C^{WP}_4$. We see that $C^{AG}_4$ would fit well with $C^{WP}_2$ too which is not surprising because the working population category is very broad (it includes 30 years, so three to four 10-year age groups).


The differences in clustering results are observed due to different categorization. The WP categorization reveals less due to less categories, however it does show the separation of two household patterns where mostly two people (couples) live,  $C^{WP}_1$ and $C^{WP}_4$. Some are still at work and the others (that sometimes live with some other family member) are already retired. AG categorization puts these two groups in the same cluster $C^{AG}_1$. We see that those 'still at work' are actually near retirement (they are about 50--60 years old) and they naturally fall into the same group. AG categorization on the other hand due to more age categories shows some more difference in the case of relationship patterns (a) \textit{respondent-parent-sibling}, $C^{AG}_2$, and (b) \textit{respondent-partner-offspring}, $C^{AG}_3$.  These two relationship patterns reveal core families with (a) a respondent being in his/hers twenties and (b) a respondent being around 30--50 years old. In these two 'types of families' $C^{AG}_2$ are about 10 to 15 years older than $C^{AG}_3$. The cluster $C^{AG}_4$ however shows additional pattern that WP categorization does not reveal -- the extended families with more females and a very specific household age pattern.

\begin{figure}
  \includegraphics[width=\textwidth]{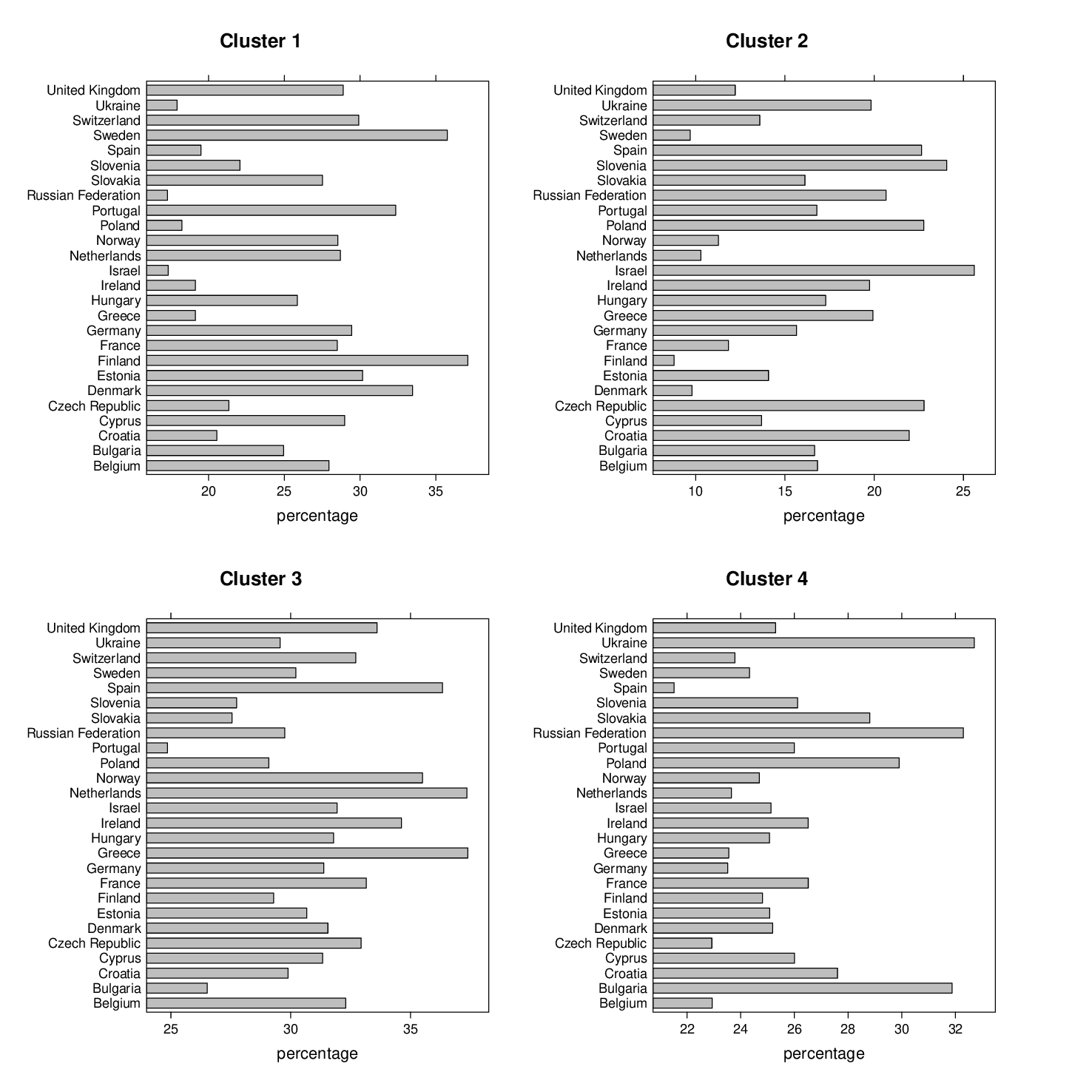}
\caption{Supplementary variable country for variable set AG.}\label{f: ctry RGY9}
\end{figure}

Considering also supplementary variable 'country' which was not included in the these two clusterings (Figure \ref{f: ctry RGY9}) we found that household pattern $C^{AG}_4$ (extended family) has the largest percentage in Ukraine, Russian Federation, Bulgaria and Poland. $C^{AG}_2$ with younger questionnaire respondent in the core family is relatively most frequent in Israel, Slovenia, Spain and Czech Republic and with older respondent, $C^{AG}_3$, in The Netherlands, Greece, Spain and Norway. Respondents living in mostly two-person families are most frequently interviewed in Finland, Sweden, Denmark and Portugal. Since ESS is one of the surveys that should represent the whole population the very large (and very small) relative values for country should exhibit a kind of household pattern that can be observed in each country (i.e. large families in Russian Federation, Spain and Ukraine).

%

%

\begin{figure}
  \includegraphics[width=\textwidth]{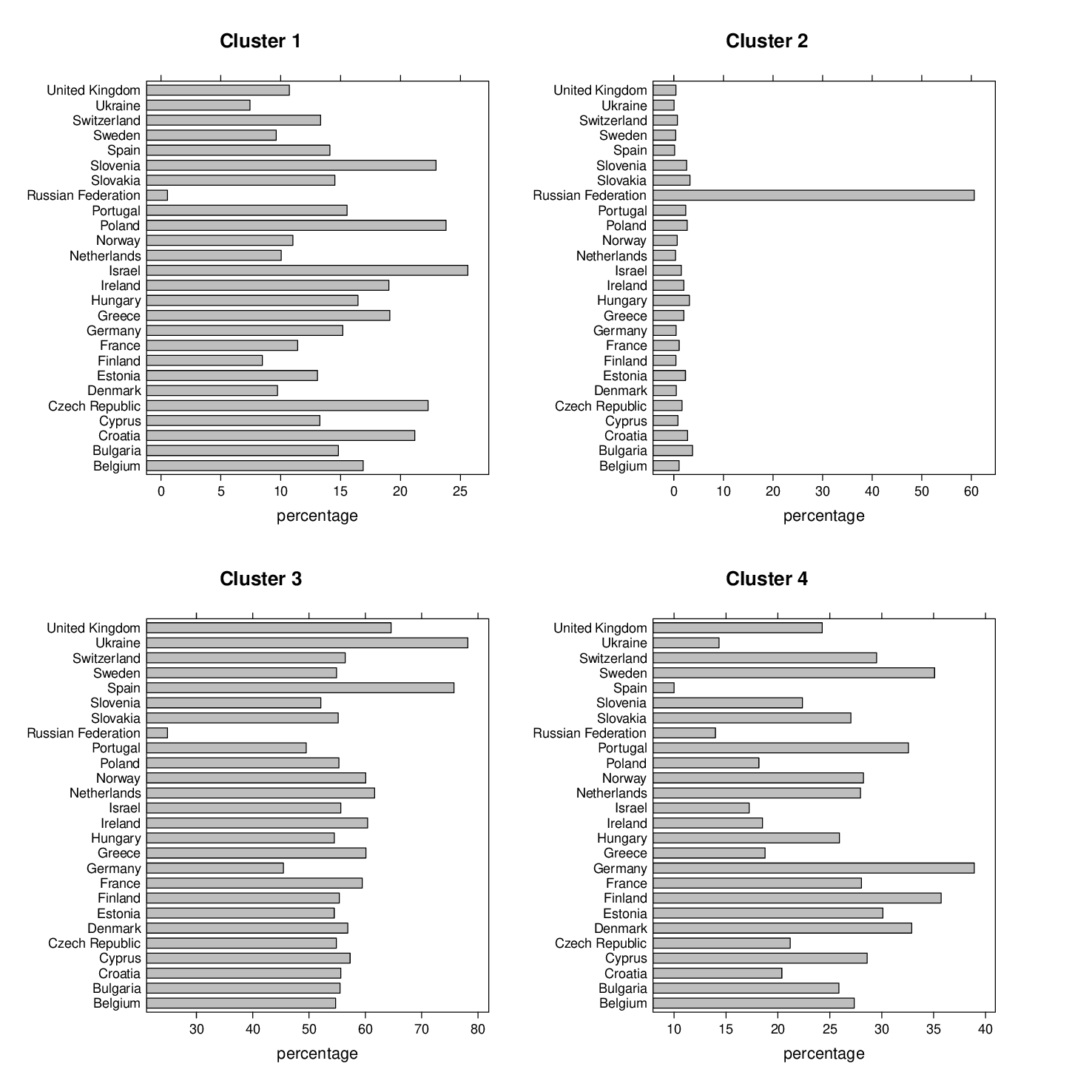}
\caption{Variable country for variable set Co with 10-year age categories.}\label{f: ctry RGYC9}
\end{figure}

These differences should be even more pronounced when 'country' is included in the clustering process. The best clustering split the data set Co (with included additional variable 'country') into one very large cluster ($C_3$ with 27,759 households), one medium sized ($C_4$ with 14,488 households) and two small clusters ($C_2$ with 2,120 and $C_1$ with 6,005 households).
Figures \ref{f: ctry RGYC9} and \ref{f: var RGYC9} show the results. Note that for easier observation scales for percentages in the horizontal coordinates are different.
 Immediately we can notice the  cluster $C_2$ with dominating  extended families in Russian Federation. This cluster is also the smallest. The largest cluster $C_3$ (core family with small proportion of other members in the household) is most evenly distributed among countries, but most pronounced in Ukraine and Spain. Shares in the second smallest cluster $C_1$  with core families and younger respondent are still large in Israel, Slovenia, Czech Republic but now also for Poland and Croatia. We could conjecture that in these countries offspring stay with parents long before becoming independent. The cluster $C_4$ belongs to older two- to three-person families with large proportions of German, Finnish, Swedish and Danish households. This type of households is the least evenly distributed among countries.

%
%

\section{Conclusion}

In the paper versions of well known leaders nonhierarchical and Ward's hierarchical methods, adapted for modal valued
symbolic data, are presented. Since the data measured in traditional measurement scales (numerical, ordinal, categorical) can all be transformed into modal symbolic representation the methods can be used for clustering data sets of mixed units.
Our approach allows the user to consider, using the weights, also the original frequency information. 
The proposed clustering methods are compatible -- they solve the same optimization problem and can be used each one separately or in combination (usually for large data sets). The optimization criterion function depends on a basic dissimilarity $\delta$ that enables user to specify different criteria. In principle, because of the additivity of components of
criterion function, we could use different $\delta$s for different symbolic variables.

Presented methods were applied on the example of household structures from the ESS 2010 data set. The clustering was done on nominal (gender, relationships, country) and interval data (age groups).
When clustering such data information on size (which is important when design and population weights have to be used to get the sample representative of a population)
was included into the clustering process.

The proposed approach is partially implemented in the R-package
\textbf{clamix} \citep{Clamix}.



\section*{Appendix}


In this Appendix we present derivations of entries $t^*_C$,  $z$ and
$D(C_u,C_v)$  from Table~\ref{t:relative} for different
basic dissimilarities $\delta$.

Since in our approach the clustering criterion function $P(\cling)$ is additive and the dissimilarity $d$ is defined component-wise, all derivations
can be limited only to a single variable and a single its component. Therefore,
the subscripts $i$ (of a variable) and $j$ (of a component) will be omitted from the expressions.

In derivations we are following the same steps as we used for $\delta_1$ in
Section~2. To obtain the component $t^*_C$ of representative of cluster $C$ we
solve for a selected $\delta$ the one dimensional optimization problem (\ref{leaderopt}). Let us denote its criterion function with $F(t), t\geq 0$
\[ F(t) = \sum_{X \in C} w_{x} \delta(p_{x},t) \]
then the optimal solution is obtained as solution of the equation 
\[ \frac{\mbox{d}\, F(t)}{\mbox{d}\, t} = 0 .\]

To obtain the component $z$ of the leader of a cluster $C_z$ we
use the relation for $t^*_C$ for a cluster $C_z$ and re-express it in terms
of quantities for $C_u$ and $C_v$.

Finally, to determine the between cluster dissimilarity $D(C_u,C_v)$ for a selected
$\delta$ we will use the relation (\ref{eqDuv}) and following the scheme for
$\delta_1$ the auxiliary quantity $S_u$ from (\ref{Su}) (after omitting indices
$i$ and $j$)

\begin{equation}\label{Suij}
 S_u  = \sum_{X \in C_u}  w_x (\delta(p_x,z) - \delta(p_x,u)) 
\end{equation}

For different combinations of the weights used in the expressions, the abbreviations $w_C, P_C, Q_C, H_C$ and $G_C$
from Table~\ref{t:relative} are used.
Note that for $C_z = C_u \cup C_v$ and $C_u \cap C_v = \emptyset$, we have
$R_{z} = R_{u}+R_{v}$, for $R \in \{ w, P, Q, H, G \}$.

\subsection{$\delta_2(x,t) = \left (\frac{p_{x}-t}{t} \right )^2$ }

\noindent \textbf{The derivation of the leader $t_C^*$:}
In this case
\[ F(t) = \sum_{X \in C} w_{x}\left (\frac{p_{x}-t}{t} \right )^2 \quad
\mbox{and} \quad
 F'(t) = -2 \sum_{X \in C} w_x p_x (p_x - t) \frac{1}{t^3} = 0 . \]
The leader's component is determined with
\begin{equation} \label{eqtC2}
t_C^* = \frac{\sum_{X \in C} w_x p_x^2}{\sum_{X \in C} w_x p_x} = \frac{Q_C}{P_C}.
\end{equation}
\noindent \textbf{The derivation of the leader $z$ of the merged disjoint clusters $C_u$ and $C_v$:}
From the Eq. (\ref{eqtC2}) and $C_u \cap C_v = \emptyset$ follows
\[ z = \frac{Q_z}{P_z} = \frac{Q_u + Q_v}{P_u + P_v} = \frac{P_u u + P_v v}{P_u + P_v} . \]
\noindent \textbf{The derivation of the dissimilarity $D(C_u,C_v)$ between the  disjoint clusters $C_u$ and $C_v$:}
Since $u$ is the leader of the cluster $C_u$ it holds $Q_u = P_u u$ (see Eq. (\ref{eqtC2})). We can replace
$\sum_{X \in C_u} w_x p_x^2 = Q_u$ in the expression $S_u$ (Eq. (\ref{Suij})) 
\[ S_u = \sum_{X \in C_u} w_{x} \left[ \left (\frac{p_{x}-z}{z} \right )^2 - \left (\frac{p_{x}-u}{u} \right )^2 \right ] \]
with $P_u u$ and get
\[ S_u = P_u \frac{(u-z)^2}{uz^2} = \frac{P_{u}}{u} \left(\frac{u-z}{z}\right)^2. \]
Similary $\displaystyle S_v = \frac{P_{v}}{v} \left(\frac{v-z}{z}\right)^2$. Combining both expressions (Eq. (\ref{Suv})) we get
\[ D(C_u,C_v) = \frac{P_{u}}{u} \left(\frac{u-z}{z}\right)^2 + \frac{P_{v}}{v} \left(\frac{v-z}{z}\right)^2. \]


\subsection{$\delta_3(x,t) = \frac{(p_{x}-t)^2}{t}$ }

\noindent \textbf{The derivation of the leader $t_C^*$:}
In this case
\[ F(t) = \sum_{X \in C} w_{x}\frac{(p_{x}-t)^2}{t}  \quad
\mbox{and} \quad
 F'(t) = \sum_{X \in C} (t^2 - p_x^2) \frac{1}{t^2} = 0 . \]
The square of the leader's component is determined with
\begin{equation} \label{eqtC3}
{t_C^*}^2 = \frac{\sum_{X \in C} w_x p_x^2}{\sum_{X \in C} w_x} = \frac{Q_C}{w_C}
\end{equation}
and from it
\[
t_C^* = \sqrt{\frac{Q_C}{w_C}} .
\]
\noindent \textbf{The derivation of the leader $z$ of the merged disjoint clusters $C_u$ and $C_v$:}
From  Eq. (\ref{eqtC3}) and $C_u \cap C_v = \emptyset$ follows
\[ z^2 = \frac{Q_z}{w_z} = \frac{Q_u + Q_v}{w_u + w_v} = \frac{w_u u^2 + w_v v^2}{w_u + w_v}  \quad
\mbox{and}
\quad  z = \sqrt{\frac{w_u u^2 + w_v v^2}{w_u + w_v}} .  \]
\noindent \textbf{The derivation of the dissimilarity $D(C_u,C_v)$ between the  disjoint clusters $C_u$ and $C_v$:}
Since $u$ is the leader of the cluster $C_u$ and $Q_u = w_u u^2$ (see Eq. (\ref{eqtC3})), we can replace
$\sum_{X \in C_u} w_x p_x^2 = Q_u$ in the expression $S_u$ 
\[ S_u = \sum_{X \in C_u} w_{x} \left [ \frac{(p_{x}-z)^2}{z} - \frac{(p_{x}-u)^2}{u} \right ] \]
with $w_u u^2$ and get
\[ S_u = w_u \frac{(u-z)^2}{z} . \]
Similary $\displaystyle S_v = w_v \frac{(v-z)^2}{z}$. Combining both expressions we get
\[ D(C_u,C_v) = w_u \frac{(u-z)^2}{z} + w_v \frac{(v-z)^2}{z} . \]

\subsection{$\delta_4(x,t) = \left (\frac{p_{x}-t}{p_{x}} \right )^2$ }

\noindent \textbf{The derivation of the leader $t_C^*$:}
In this case
\[ F(t) = \sum_{X \in C} w_{x}\left (\frac{p_{x}-t}{p_{x}} \right )^2  \quad
\mbox{and} \quad
 F'(t) = -2 \sum_{X \in C} w_x (p_x - t) \frac{1}{p_x^2} = 0 . \]
For $p_x \neq 0$ (this is also the condition for $\delta_4(x,t)$ to be defined), the leader's component is determined with
\begin{equation} \label{eqtC4}
t_C^* = \frac{\sum_{X \in C} \frac{w_x}{p_x}}{\sum_{X \in C} \frac{w_x}{p_x^2}} = \frac{H_C}{G_C}.
\end{equation}
In the case $p_x = 0$ we set $t_C^* = 0$ and $\delta_4(x,t) = 0$.

\noindent \textbf{The derivation of the leader $z$ of the merged disjoint clusters $C_u$ and $C_v$:}
From Eq. (\ref{eqtC4}) and $C_u \cap C_v = \emptyset$ follows
\[ z = \frac{H_z}{G_z} = \frac{H_u + H_v}{G_u + G_v} = \frac{H_{u}+H_{v}}{\frac{H_{u}}{u} + \frac{H_{v}}{v}} . \]
\noindent \textbf{The derivation of the dissimilarity $D(C_u,C_v)$ between the  disjoint clusters $C_u$ and $C_v$:}
Since $u$ is the leader of the cluster $C_u$ and $H_u = G_u u$ (see Eq. (\ref{eqtC4})), we can replace
$\sum_{X \in C_u} \frac{w_x}{p_x} = H_u$ in the expression $S_u$ 
\[ S_u = \sum_{X \in C_u} w_{x} \left[ \left (\frac{p_{x}-z}{p_{x}} \right )^2 - \left (\frac{p_{x}-u}{p_{x}} \right )^2 \right ] \]
with $G_u u$ and get
\[ S_u = G_u (u-z)^2 . \]
Similary $\displaystyle S_v = G_v (v-z)^2$. Combining both expressions we get
\[ D(C_u,C_v) = G_u (u-z)^2 + G_v (v-z)^2 . \]

\subsection{$\delta_5(x,t) = \frac{(p_{x}-t)^2}{p_{x}}$ }

\noindent \textbf{The derivation of the leader $t_C^*$:}
In this case
\[ F(t) = \sum_{X \in C} w_{x} \frac{(p_{x}-t)^2}{p_{x}}  \quad
\mbox{and} \quad
 F'(t) = -2 \sum_{X \in C} w_x (p_x - t) \frac{1}{p_x} = 0 . \]
For $p_x \neq 0$ (this is also the condition for $\delta_5(x,t)$ to be defined), the leader's component is determined with
\begin{equation} \label{eqtC5}
t_C^* = \frac{\sum_{X \in C} w_x}{\sum_{X \in C} \frac{w_x}{p_x}} = \frac{w_C}{H_C}.
\end{equation}
In the case $p_x = 0$ we set $t_C^* = 0$ and $\delta_5(x,t) = 0$.

\noindent \textbf{The derivation of the leader $z$ of the merged disjoint clusters $C_u$ and $C_v$:}
From Eq. (\ref{eqtC5}) and $C_u \cap C_v = \emptyset$ follows
\[ z = \frac{w_z}{H_z} = \frac{w_u + w_v}{H_u + H_v}  . \]
\noindent \textbf{The derivation of the dissimilarity $D(C_u,C_v)$ between the  disjoint clusters $C_u$ and $C_v$:}
Since $u$ is the leader of the cluster $C_u$ and $w_u = H_u u$ (see Eq. (\ref{eqtC5})), we can replace
$\sum_{X \in C_u} w_x = w_u$ in the expression $S_u$ 
\[ S_u = \sum_{X \in C_u} w_{x} \left[ \frac{(p_{x}-z)^2}{p_{x}} - \frac{(p_{x}-u)^2}{p_{x}} \right ] \]
with $H_u u$ and get
\[ S_u = H_u (u-z)^2 = w_u \frac{(u-z)^2}{u}. \]
Similary $\displaystyle S_v = w_v \frac{(v-z)^2}{v}$. Combining both expressions we get
\[ D(C_u,C_v) = w_u \frac{(u-z)^2}{u} + w_v \frac{(v-z)^2}{v} . \]

\subsection{$\delta_6(x,t) = \frac{(p_{x}-t)^2}{p_{x} \cdot t}$ }

\noindent \textbf{The derivation of the leader $t_C^*$:}
In this case
\[ F(t) = \sum_{X \in C} w_{x} \frac{(p_{x}-t)^2}{p_{x}\cdot t}  \quad
\mbox{and} \quad
 F'(t) = \sum_{X \in C} \frac{w_x}{p_x} (t^2 - p_x^2) = 0 . \]
For $p_x \neq 0$ (this is also the condition for $\delta_6(x,t)$ to be defined), the  leader's component is determined with
\begin{equation} \label{eqtC6}
{t_C^*}^2 = \frac{\sum_{X \in C} w_x p_x}{\sum_{X \in C} \frac{w_x}{p_x}} = \frac{P_C}{H_C}
\end{equation}
and from it
\[
t_C^* = \sqrt{\frac{P_C}{H_C}} .
\]
In the case $p_x = 0$ we set $t_C^* = 0$ and $\delta_6(x,t) = 0$.

\noindent \textbf{The derivation of the leader $z$ of the merged disjoint clusters $C_u$ and $C_v$:}
From Eq. (\ref{eqtC6}) and $C_u \cap C_v = \emptyset$ follows
\[ z^2 = \frac{P_z}{H_z} = \frac{P_u + P_v}{H_u + H_v} =  \frac{P_{u}+P_{v}}{\frac{P_{u}}{u^2} + \frac{P_{v}}{v^2}} \quad
\mbox{and}
\quad z = \sqrt{\frac{P_{u}+P_{v}}{\frac{P_{u}}{u^2} + \frac{P_{v}}{v^2}}} . \]
\noindent \textbf{The derivation of the dissimilarity $D(C_u,C_v)$ between the  disjoint clusters $C_u$ and $C_v$:}
Since $u$ is the leader of the cluster $C_u$ and $P_u = H_u u^2$ (see Eq. (\ref{eqtC6})), we can replace
$\sum_{X \in C_u} w_x p_x = P_u$ in the expression $S_u$ 
\[ S_u = \sum_{X \in C_u} w_{x} \left[ \frac{(p_{x}-z)^2}{p_{x}\cdot z} - \frac{(p_{x}-u)^2}{p_{x}\cdot u} \right] \]
with $H_u u^2$ and get
\[ S_u = P_u \frac{(u-z)^2}{u^2z} = \frac{P_{u}}{u} \frac{(u-z)^2}{uz}. \]
Similary $\displaystyle S_v = \frac{P_{v}}{v} \frac{(v-z)^2}{vz}$. Combining both expressions we get
\[ D(C_u,C_v) = \frac{P_{u}}{u} \frac{(u-z)^2}{uz} + \frac{P_{v}}{v} \frac{(v-z)^2}{vz} . \]

\end{document}